\documentclass[journal]{IEEEtran}

\usepackage{amsmath,amsfonts}
\usepackage{graphicx}
\usepackage{booktabs}
\usepackage{cite}
\usepackage{multirow}
\usepackage{xcolor}
\usepackage{url}

\newcommand{\dataset}{InstaBind-Lite}
\newcommand{\problem}{Dense Same-Class Attribute Misbinding}
\newcommand{\mbr}{\mathrm{MBR}}
\newcommand{\ambr}{\mathrm{A\mbox{-}MBR}}

\begin{document}

\title{\dataset: Diagnosing Dense Same-Class Attribute Misbinding in Large Vision-Language Models}

\author{Yuanzhi Xu, Qian Gao, Jun Fan, Guohui Ding, Zhenyu Yang,
        Yuteng Xiao, and Sixue Lin%
\thanks{Yuanzhi Xu, Qian Gao, Zhenyu Yang, and Yuteng Xiao are with
Qilu University of Technology (Shandong Academy of Sciences), Jinan, China
(e-mail: 10431250244@stu.qlu.edu.cn, gq@qlu.edu.cn,
yzy@qlu.edu.cn, and yutengxiao@qlu.edu.cn). Qian Gao is the corresponding author.}%
\thanks{Jun Fan is with China Telecom Digital Intelligence Technology Co., Ltd.,
Jinan, China (e-mail: fanjun.sd@chinatelecom.cn).}%
\thanks{Guohui Ding is with Shenyang Aerospace University, Shenyang, China
(e-mail: dingguohui@sau.edu.cn).}%
\thanks{Sixue Lin is with University of Nottingham Ningbo China, Ningbo, China
(e-mail: scxsl5@nottingham.edu.cn).}
}

\maketitle

\begin{abstract}
Large vision-language models can recognize the objects and attributes in a crowded scene yet assign an attribute to the wrong same-class instance. Generic visual-question-answering accuracy marks the response as wrong, while object-hallucination metrics may regard both the object and attribute as image-supported; neither reveals the transfer. This study formalizes this blind spot as \emph{\problem} and presents \dataset, a controlled benchmark that makes it directly measurable. Its 524 images contain 529 curated groups of 3--6 same-class entities, 1773 boxed instances, ordered neighbors, distinguishable color-like attributes, and four complementary question levels, yielding 9580 deterministically evaluated questions. Unlike existing protocols, source-instance annotations separate unsupported generation and recognition failure from an attribute copied from another visible entity. Binding-specific metrics further quantify transfer frequency, adjacency, ordinal distance, and intervention effects. Across five open-source and two commercial/API models, the open-source systems average 19.84\% Misbinding Rate and the API systems 7.55\%; these errors are hidden by aggregate accuracy. Among identifiable transfers, 80.70\% and 81.51\%, respectively, originate from adjacent instances. Localization and instance-first interventions help selected models but are not universal remedies. \dataset therefore turns previously undifferentiated wrong answers into source-identifiable failure categories and tests a reliability dimension that conventional benchmarks cannot determine: whether a model knows not only \emph{what} is visible, but \emph{which instance} owns each attribute.
\end{abstract}


\section{Introduction}

Large vision-language models (LVLMs) now support image search, visual assistance, traffic analysis, retail inspection, and robotic interaction~\cite{radford2021clip,li2022blip,alayrac2022flamingo,li2023blip2,dai2023instructblip}. Their errors, however, have different causes and risks: a model may invent absent content, miss an object, misread an attribute or relation, or attach a correctly perceived attribute to the wrong entity. Existing benchmarks often collapse these outcomes into the same incorrect answer, so improved aggregate accuracy does not establish reliable instance--attribute correspondence~\cite{antol2015vqa,li2023pope,wang2023amber,guan2024hallusionbench}. A correct color attached to the wrong vehicle, person, product, or container can still produce unsafe entity-specific retrieval or action.

Public incidents outside LVLM benchmarking illustrate the stakes of visual attribution failure. A traffic-recognition system in Ningbo reportedly treated a face printed on a bus advertisement as a jaywalking pedestrian~\cite{guardian2018busadvert}; the U.S. National Transportation Safety Board found that the developmental automated-driving system in the 2018 Tempe crash repeatedly changed its classification of a pedestrian before the fatal collision~\cite{ntsb2019tempe}. Neither incident establishes DSCAM or evaluates an LVLM, but both show why visually supported content is insufficient when identity, location, and attributes must remain attached to the correct entity. The analogous LVLM error can direct a user to the wrong person, product, vehicle, or waste stream.

Generic visual-question-answering benchmarks collapse these cases into the same zero-accuracy event, whereas object-hallucination benchmarks mainly test whether predicted content is supported somewhere in the image~\cite{rohrbach2018object,li2023pope,wang2023amber,guan2024hallusionbench}. They therefore cannot determine whether a multimedia system preserves \emph{which entity has which property}.

This study defines the hidden failure as \emph{Dense Same-Class Attribute Misbinding} (DSCAM): a wrong answer copied from another visible same-class instance. The hypothesis is structural: under same-class competition, transfers should disproportionately originate from neighbors rather than uniformly random sources. Identifying that source separates binding failure from recognition error and gives grounding, decoding, and training methods a specific repair target.

Existing resources lack the joint annotations needed to test this hypothesis. General VQA does not guarantee inspectable same-class groups with distinguishable attributes; grounding evaluates localization but not which neighbor supplied a wrong attribute; and compositionality benchmarks do not measure directed instance-to-instance transfer in natural dense scenes. Their standard scores therefore cannot recover DSCAM prevalence, source, or distance.

\dataset supplies that missing capability through ordered same-class groups, visible color-like attributes, boxes, and neighbor links. Four levels probe position-to-attribute, attribute-to-position, relation-mediated binding, and proposition verification. A controlled vocabulary and source annotations support deterministic parsing into adjacent or non-adjacent misbinding, recognition failure, and out-of-set hallucination. Across seven LVLMs, this design reveals measurable MBR even at high aggregate accuracy and shows that transfers overwhelmingly follow neighbor structure.

The contributions are threefold:
\begin{itemize}
    \item \problem is formalized as a distinct failure category, resolving the evaluation ambiguity between unsupported hallucination, attribute recognition failure, and a visible attribute transferred to the wrong same-class instance.
    \item \dataset is constructed as a high-purity diagnostic benchmark with 524 images, 1773 instances, and 9580 questions. Its controlled groups and source-instance annotations make wrong-instance binding measurable in a way that aggregate VQA and object-presence scores do not.
    \item Binding-oriented metrics and controlled interventions are introduced. MBR identifies visible wrong-instance transfers, A-MBR and ordinal Distance-MBR locate their spatial source, while crop/context localization and instance-first prompting test whether reducing visual competition or explicitly enumerating instance--attribute pairs mitigates the failure.
\end{itemize}

\section{Related Work}

Table~\ref{tab:benchmark_capability} separates benchmark breadth from diagnostic observability. The comparison does not imply that a focused benchmark is generally superior to broad evaluation; it identifies which protocols can trace a wrong attribute to a competing same-class source instance.

\begin{table*}[t]
\centering
\caption{Comparison by diagnostic objective. ``Partial'' indicates that same-class instances may occur naturally, but are not jointly controlled with source-instance attribution for misbinding analysis.}
\label{tab:benchmark_capability}
\scriptsize
\renewcommand{\arraystretch}{1.12}
\setlength{\tabcolsep}{3pt}
\begin{tabular}{p{0.15\textwidth} p{0.21\textwidth} p{0.17\textwidth} p{0.16\textwidth} p{0.16\textwidth}}
\toprule
Evaluation family & Primary diagnostic target & Controlled same-class groups & Wrong-source attribution & Spatial misbinding metrics \\
\midrule
Object hallucination & Unsupported objects or attributes & No & No & No \\
General VQA / multimodal & End-answer correctness and broad capability & Partial & No & No \\
Referring / grounding & Phrase-to-region localization & Partial & No & No \\
Compositionality & Attribute--object and relation sensitivity & Partial & No & No \\
\dataset & Same-class instance--attribute binding & Yes & Yes & Yes \\
\bottomrule
\end{tabular}
\end{table*}


\subsection{LVLM Hallucination Evaluation}
CHAIR identifies absent objects in captions~\cite{rohrbach2018object}; POPE tests object existence~\cite{li2023pope}; AMBER provides judge-free multidimensional analysis~\cite{wang2023amber}; and HallusionBench couples language hallucination with visual illusion~\cite{guan2024hallusionbench}. Detection and decoding methods likewise target unsupported generation~\cite{gunjal2024mhaldetect,huang2024opera}. DSCAM is different: both class and attribute are image-supported, but their instance correspondence is false. Presence-based protocols either miss this event or count it only as an unspecified error.

\subsection{Visual Question Answering and General Multimodal Benchmarks}
VQA, VQA v2, and Visual7W evaluate image-conditioned answers and reduce language bias or add region grounding~\cite{antol2015vqa,goyal2017vqav2,zhu2016visual7w}. COCO, Visual Genome, GQA, and VAW provide categories, boxes, scene graphs, relations, and attributes~\cite{lin2014coco,krishna2017visualgenome,hudson2019gqa,pham2021vaw}; MME, SEED-Bench, MM-Vet, and MMBench broaden capability coverage~\cite{fu2023mme,li2023seedbench,yu2024mmvet,liu2024mmbench}. Their standard protocols do not jointly control same-class density, attribute uniqueness, ordered neighbors, and wrong-source attribution. An adjacent transfer thus receives the same score as an absent-color guess. \dataset trades breadth for controlled evidence that separates these risks.

\subsection{Visual Grounding and Referring Expressions}
Flickr30K Entities, ReferItGame, and RefCOCO associate expressions with image regions, while MAttNet decomposes references into subject, location, and relation modules~\cite{plummer2017flickr30k,kazemzadeh2014referitgame,yu2016referringcontext,yu2018mattnet}. Their output is typically a region scored by localization overlap. DSCAM instead tests attribution after or alongside selection: a model may locate the target yet answer with its neighbor's attribute. \dataset therefore retains boxes for interventions but adds ordered attributes and source-instance error labels that overlap alone cannot provide.

\subsection{Compositionality and Attribute--Relation Binding}
CLEVR studies compositional reasoning in synthetic scenes~\cite{johnson2017clevr}; Winoground contrasts image--caption pairs with shared words but different compositions~\cite{thrush2022winoground}; and ARO, VL-CheckList, CREPE, and SugarCrepe test attributes, relations, and word order~\cite{yuksekgonul2023bags,zhao2022vlchecklist,ma2023crepe,hsieh2023sugarcrepe}. Winoground's minimal-pair principle motivates separating component recognition from composition in \dataset: the same-class set and vocabulary remain fixed while the queried instance changes. ARO's relation sensitivity motivates L3 and ordinal-distance analysis. Thus, natural-image composition becomes a directed transfer from target $i$ to source $j$, and MBR connects failure to a competing entity rather than only an incorrect global match.

\subsection{Binding Mechanisms and Multi-Subject Misbinding}
Mechanistic studies examine object--reference binding and its distribution across vision encoders and language backbones~\cite{saravanan2025binding,cui2026dualbinding}; MultiBind studies cross-subject attribute transfer in image generation~\cite{tian2026multibind}. This work asks the complementary behavioral question of whether an LVLM reading a natural image assigns an observed attribute to the correct same-class instance. Spatial signatures and interventions can guide, but do not replace, mechanistic analysis.

\section{Problem Definition}

Let an image contain a same-class group
\begin{equation}
G = \{e_1, e_2, \ldots, e_n\},
\end{equation}
where all entities share class $c$ and have a spatial order, e.g., left-to-right. Each entity $e_i$ has an attribute value $a_i$ for an attribute type such as color or upper-clothing color.

A question specifies a target entity $e_t$ directly or indirectly. In L1, the target is specified by position and the answer is its attribute. In L2, the attribute is specified and the answer is the target position. In L3, the target is obtained through a local relation such as left or right neighbor. In L4, the model verifies an instance-attribute proposition. A model prediction is correct if it matches the canonical answer after answer normalization.

If the prediction is incorrect but matches the attribute or position of another same-class entity $e_j$, the error is classified as a misbinding:
\begin{equation}
\hat{a} = a_j,\quad j \ne t.
\end{equation}

The instance distance of a misbinding is defined by the absolute order difference:
\begin{equation}
d(t,j) = | \mathrm{order}(e_t) - \mathrm{order}(e_j) |.
\end{equation}
An adjacent misbinding occurs when $d(t,j)=1$.

This definition intentionally separates misbinding from out-of-set hallucination. If a model predicts an attribute that does not appear in the same-class group, the error is counted as out-of-set hallucination rather than misbinding. This distinction tests whether a wrong answer is grounded in the image but attached to the wrong instance.

The key diagnostic question is therefore not only whether the model answers correctly, but what kind of wrong answer it gives. A random attribute error, an out-of-image hallucination, and a neighbor's attribute are all incorrect under accuracy, but they imply different failure mechanisms. \problem focuses on the last case.

\section{\dataset Dataset}

\begin{table}[t]
\centering
\caption{Dataset statistics of InstaBind-Lite v0.4.}
\label{tab:dataset_stats}
\footnotesize
\setlength{\tabcolsep}{4pt}
\begin{tabular}{lr}
\toprule
Item & Value \\
\midrule
Images & 524 \\
Same-class groups & 529 \\
Instances & 1773 \\
Questions & 9580 \\
Person groups & 36.86\% \\
Non-person groups & 63.14\% \\
Group size 3/4/5/6 & 391 / 98 / 32 / 8 \\
Question L1/L2/L3/L4 & 1773 / 1773 / 2488 / 3546 \\
Sources: COCO/GQA/VAW & 205 / 21 / 4 \\
Sources: web/self-shot & 277 / 17 \\
\bottomrule
\end{tabular}
\end{table}


\dataset is a high-purity diagnostic benchmark rather than a general-purpose VQA dataset. Each image is curated to contain 3--6 same-class entities with clear order, visible attributes, and low ambiguity, making targets and neighbors human-inspectable. Table~\ref{tab:dataset_stats} summarizes the 524 images, 529 groups, 1773 instances, and 9580 questions.

\subsection{Data Sources}
Images come from COCO~\cite{lin2014coco}, GQA~\cite{hudson2019gqa}, VAW~\cite{pham2021vaw}, manually selected open-license web images, and self-shot images. Manual verification removes incomplete groups, severe occlusion, strong reflection, tiny objects, and uncertain attributes. Source metadata and original licenses remain separate from benchmark annotations.

\subsection{Annotation Schema}
Each group records its image, class, spatial order, boxes, attributes, and neighbor links. Boxes support quality control and interventions; questions use natural-language positions and relations. Person questions use upper-clothing color, and transparent or multicolor labels are retained only when unambiguous.

\subsection{Attribute Scope and Design Trade-off}
The benchmark focuses on color and upper-clothing color: local attributes shared across classes and expressible with a compact vocabulary. This enables source attribution without an LLM judge; L2 additionally requires within-group attribute uniqueness. The scope is a controlled slice of binding and does not assume unchanged rates for actions, materials, textures, shapes, or states.

\subsection{Question Levels}
The benchmark generates four levels of questions:
\begin{itemize}
    \item L1: position-to-attribute, e.g., ``What color is the leftmost car?''
    \item L2: attribute-to-position, e.g., ``Where is the red car?''
    \item L3: relation-interference, e.g., ``What color is the car to the right of the red car?''
    \item L4: instance verification, e.g., ``Is the middle car red?''
\end{itemize}
L2 questions require the queried attribute to be unique within the same-class group.

Together, L1 tests position-conditioned reading, L2 reverse binding, L3 local relational interference, and L4 proposition verification, separating perception errors from instance-level transfer.

\section{Metrics}

\subsection{Accuracy}
Accuracy measures whether the normalized model prediction matches the canonical answer. It provides the standard task-level score but does not explain how wrong answers relate to the image content.

\subsection{Misbinding and Adjacency}
Let $\mathcal{Q}$, $\mathcal{W}$, $\mathcal{M}$, and $\mathcal{A}$ denote all questions, wrong answers, misbindings, and adjacent misbindings. The three rates are
\begin{equation}
\mbr=\frac{|\mathcal{M}|}{|\mathcal{Q}|},\quad
\mathrm{Err\mbox{-}MBR}=\frac{|\mathcal{M}|}{|\mathcal{W}|},\quad
\ambr=\frac{|\mathcal{A}|}{|\mathcal{M}|}.
\end{equation}
MBR measures transfer frequency, error-conditioned MBR its share among wrong answers, and A-MBR whether transfers follow within-group neighbor structure.

\subsection{Out-of-Set Hallucination}
Out-of-set hallucination measures predictions that do not match any same-class instance attribute or position. This separates ungrounded attributes from in-image but misbound attributes. A model can therefore have a low object hallucination profile while still exhibiting substantial instance-level misbinding.

\subsection{Distance-MBR and Confusion Matrix}
Distance-MBR analyzes the distribution of misbinding over the ordinal distance $d(t,j)$ defined in Section~III. Thus, distance 1 denotes adjacent instances in the annotated left-to-right order; it is not a Euclidean pixel-distance bin. This choice is robust to image scale and uneven object spacing and directly tests local instance competition, but it cannot determine whether ordinal adjacency or physical separation is the dominant cause. Misbinding confusion matrices record how often the target instance index $i$ is confused with source instance index $j$.

\subsection{Intervention Gap}
Full-image evaluation is compared with localized variants, including crop oracle and context crop. The binding gap is the accuracy difference between an intervention setting and the full-image setting on the same underlying target-question subset. Because crop views require a localized reformulation of the query, these are joint visual-and-query interventions rather than image-only ablations. A positive gap suggests that reducing same-class visual competition helps the model, while a small or negative gap indicates that localization is insufficient or that the model relies on broader context.

\section{Experiments}

\subsection{Models and Inference Protocol}
The evaluation covers five open-source LVLMs and two commercial/API LVLMs: Qwen2.5-VL-7B~\cite{bai2025qwen25vl}, InternVL3-8B~\cite{zhu2025internvl3}, LLaVA-1.5-7B~\cite{liu2023llava,liu2023improvedllava}, LLaVA-OneVision-7B~\cite{li2024llavaonevision}, MiniCPM-V-2.6~\cite{yao2024minicpmv}, Gemini-3.5-Flash~\cite{google2026gemini35flash}, and Qwen3-VL-Plus~\cite{qwencloud2026qwen3vlplus}. Every model receives the same question set, answer constraints, and normalized parser. Deterministic or low-temperature decoding is used whenever supported, and prompts request short answers so that evaluation reflects visual binding rather than generation style.

\subsection{Main Results: What Aggregate Accuracy Conceals}
\begin{table*}[t]
\centering
\caption{Full-image performance and misbinding metrics on InstaBind-Lite.}
\label{tab:main_full_image}
\small
\setlength{\tabcolsep}{12pt}
\begin{tabular}{lrrrrrrr}
\toprule
Model & Type & Acc. & MBR & Err-MBR & A-MBR & Out-set & Invalid \\
\midrule
Qwen2.5-VL-7B & Open & 75.89 & 13.65 & 56.62 & 82.65 & 7.07 & 0.0 \\
InternVL3-8B & Open & 76.69 & 14.01 & 60.1 & 84.65 & 7.36 & 0.0 \\
LLaVA-1.5-7B & Open & 54.37 & 34.01 & 74.54 & 78.7 & 7.89 & 0.0 \\
LLaVA-OneVision-7B & Open & 72.3 & 18.58 & 67.07 & 76.97 & 6.88 & 0.0 \\
MiniCPM-V-2.6 & Open & 71.18 & 18.96 & 65.77 & 80.51 & 7.13 & 0.0 \\
Gemini-3.5-Flash & API & 80.13 & 5.79 & 29.15 & 78.38 & 9.66 & 1.42 \\
Qwen3-VL-Plus & API & 81.52 & 9.31 & 50.4 & 84.64 & 7.53 & 0.02 \\
\bottomrule
\end{tabular}
\end{table*}


Table~\ref{tab:main_full_image} demonstrates the diagnostic information added by \dataset. Aggregate accuracy identifies Qwen3-VL-Plus and Gemini-3.5-Flash as the strongest evaluated systems, at 81.52\% and 80.13\%, but it cannot state why their remaining answers are wrong. MBR reveals that 9.31\% and 5.79\% of \emph{all} questions, respectively, are answered with information belonging to another visible same-class instance. Thus, high general accuracy does not imply reliable instance--attribute correspondence.

The distinction is larger for open-source models. Their MBR ranges from 13.65\% to 34.01\%; for LLaVA-1.5, 74.54\% of all wrong answers are identifiable wrong-instance transfers rather than arbitrary mistakes. Moreover, MBR exceeds out-of-set hallucination in six of seven models. A conventional object-presence evaluation would therefore underdescribe a substantial error component: the predicted attribute often exists in the scene and is visually supported, but its ownership is wrong. These results do not negate improvements measured by general benchmarks; they show that claims of visual reliability require a complementary instance-binding test.

The second finding concerns error geometry. A-MBR ranges from 76.97\% to 84.65\% across all seven models. Even the two API systems retain A-MBRs of 78.38\% and 84.64\%, despite their lower total MBR. Stronger models reduce the frequency of misbinding, but the residual failures preserve the same local signature. This consistency across architectures and capability levels is evidence against uniform attribute guessing and supports competition between neighboring same-class representations.

\begin{figure*}[t]
    \centering
    \includegraphics[width=\textwidth]{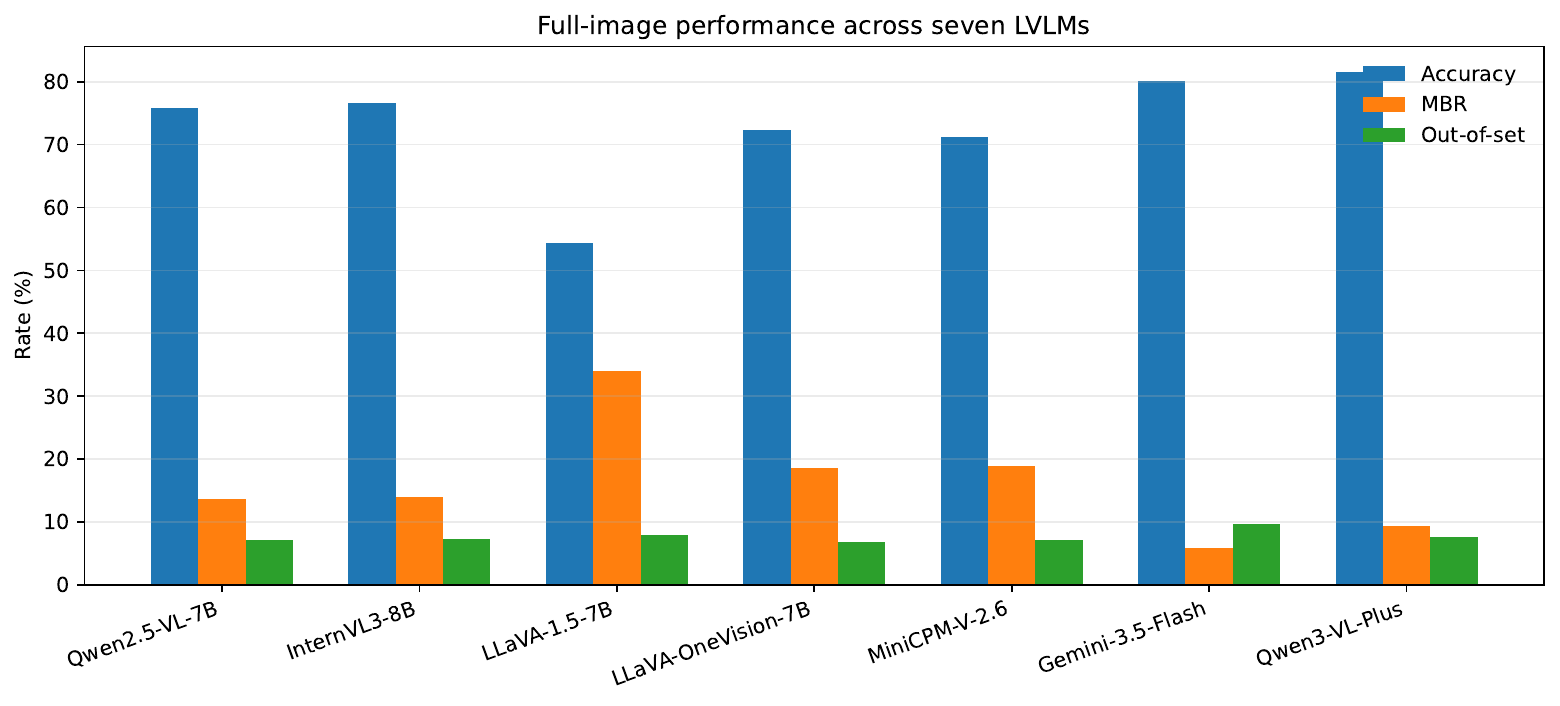}
    \caption{Full-image accuracy, MBR, and out-of-set hallucination across seven LVLMs. Aggregate accuracy and object-level hallucination do not reveal the visible wrong-instance transfers measured by MBR.}
    \label{fig:main_metrics}
\end{figure*}

\begin{figure*}[t]
    \centering
    \includegraphics[width=\textwidth]{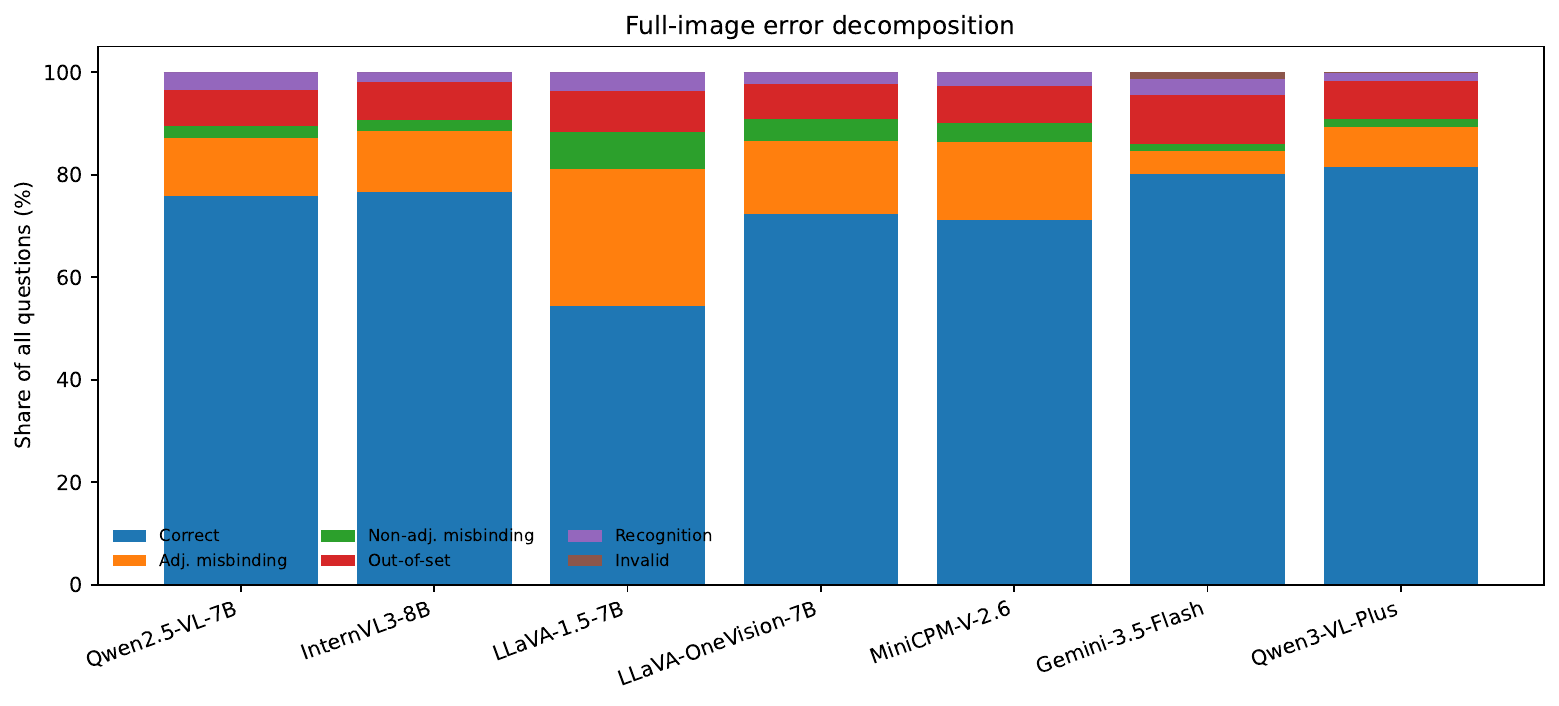}
    \caption{Full-image error decomposition. Adjacent misbinding forms a substantial error component in open-source models and remains structurally visible in API models.}
    \label{fig:error_decomposition}
\end{figure*}

\begin{figure*}[t]
    \centering
    \includegraphics[width=0.75\textwidth, keepaspectratio=true]{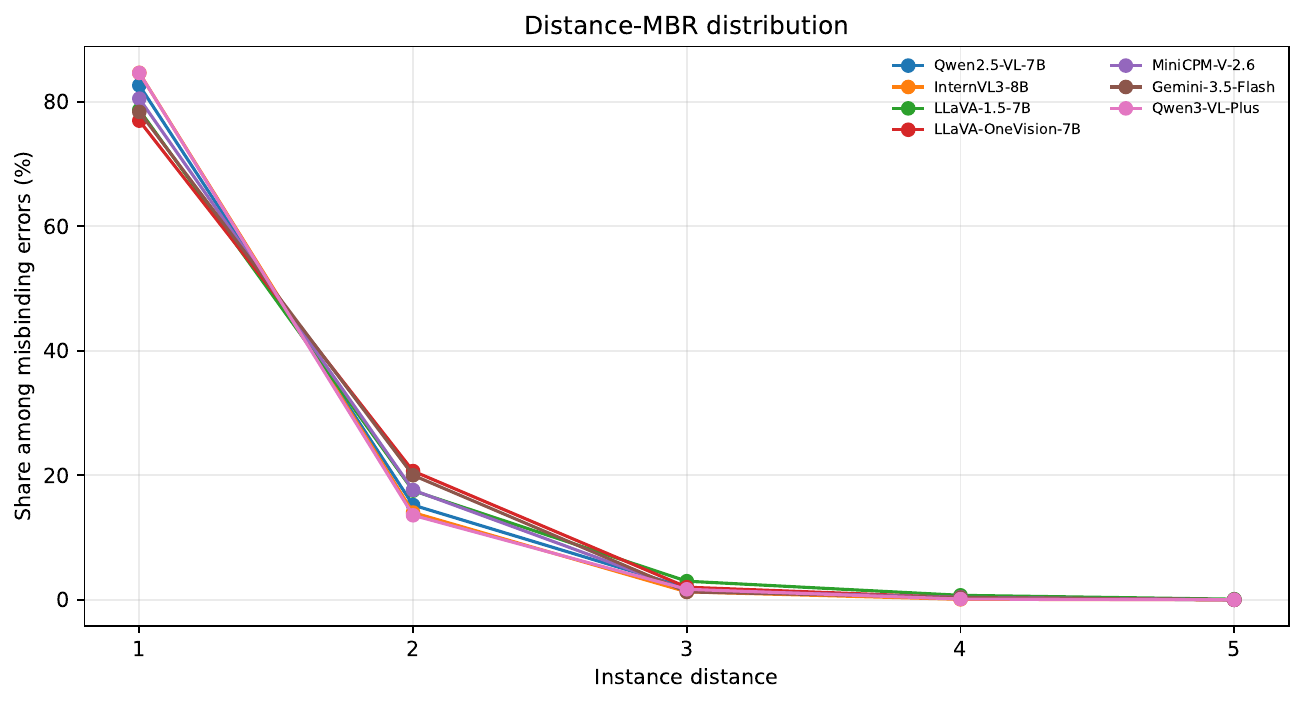}
    \caption{Ordinal Distance-MBR. Most identifiable transfers originate from the immediately adjacent same-class instance ($d=1$), while errors rapidly decline for more distant instances.}
    \label{fig:distance_mbr}
\end{figure*}

\subsection{Statistical Stability}
\begin{table}[t]
\centering
\caption{Image-cluster bootstrap 95\% confidence intervals for full-image evaluation. All intervals are computed by resampling images with replacement, preserving all questions from each sampled image.}
\label{tab:bootstrap_ci}
\scriptsize
\renewcommand{\arraystretch}{1.32}
\setlength{\tabcolsep}{1.8pt}
\begin{tabular}{lccc}
\toprule
Model & Accuracy & MBR & A-MBR \\
\midrule
Qwen2.5-VL-7B & 75.89 [74.50, 77.50] & 13.65 [12.50, 14.83] & 82.65 [80.08, 85.36] \\
InternVL3-8B & 76.69 [75.28, 78.12] & 14.01 [12.94, 15.03] & 84.65 [82.33, 86.97] \\
LLaVA-1.5-7B & 54.37 [53.05, 55.78] & 34.01 [32.70, 35.24] & 78.70 [77.15, 80.54] \\
LLaVA-OneVision-7B & 72.30 [70.80, 73.73] & 18.58 [17.41, 19.85] & 76.97 [75.03, 78.95] \\
MiniCPM-V-2.6 & 71.18 [69.69, 72.70] & 18.96 [17.77, 20.13] & 80.51 [78.66, 82.38] \\
Gemini-3.5-Flash & 80.13 [78.57, 81.54] & 5.79 [5.02, 6.53] & 78.38 [73.93, 82.85] \\
Qwen3-VL-Plus & 81.52 [80.17, 82.96] & 9.31 [8.36, 10.29] & 84.64 [81.34, 87.99] \\
\bottomrule
\end{tabular}
\end{table}


Statistical stability is evaluated with 1000 image-cluster bootstrap resamples. Images, rather than individual questions, are sampled with replacement, and all questions from each selected image remain together. This prevents the many templates derived from one scene from being treated as independent evidence. Table~\ref{tab:bootstrap_ci} shows narrow MBR intervals relative to the cross-model differences. LLaVA-1.5 remains distinctly high at 34.01\% [32.70, 35.24], whereas Gemini-3.5-Flash remains distinctly low at 5.79\% [5.02, 6.53]; neither result is explained by a small set of unusual images. More importantly, every A-MBR interval remains high: even Gemini's lower bound is 73.93\%, and Qwen3-VL-Plus reaches 84.64\% [81.34, 87.99]. Total error frequency is model-dependent, but neighbor concentration is stable under image-level sampling variation. The bootstrap therefore supports two separate conclusions: model strength changes \emph{how often} misbinding occurs, while the benchmark consistently reveals \emph{where} the transferred attribute comes from.

\subsection{Parser Reliability}
A stratified audit covers 200 outputs from the five open-source models, with 40 outputs per model. It includes 87 full-image and 113 intervention outputs, all four question levels, and five outcome strata: 50 correct answers, 50 adjacent misbindings, 35 non-adjacent misbindings, 35 out-of-set hallucinations, and 30 attribute-recognition failures. Human judgment agrees with the normalized parser on all 200 outputs. The point estimate is therefore 100\%, with a 95\% Wilson interval of approximately [98.1\%, 100\%]. This finite audit does not prove that parsing is error-free outside the sample, but it rules out a parser disagreement rate large enough to explain the reported MBR and A-MBR patterns and verifies that the error taxonomy can be applied reproducibly across answer formats and intervention settings.

\section{Qualitative Analysis and Practical Significance}

\begin{figure*}[t]
\centering
\includegraphics[width=\textwidth]{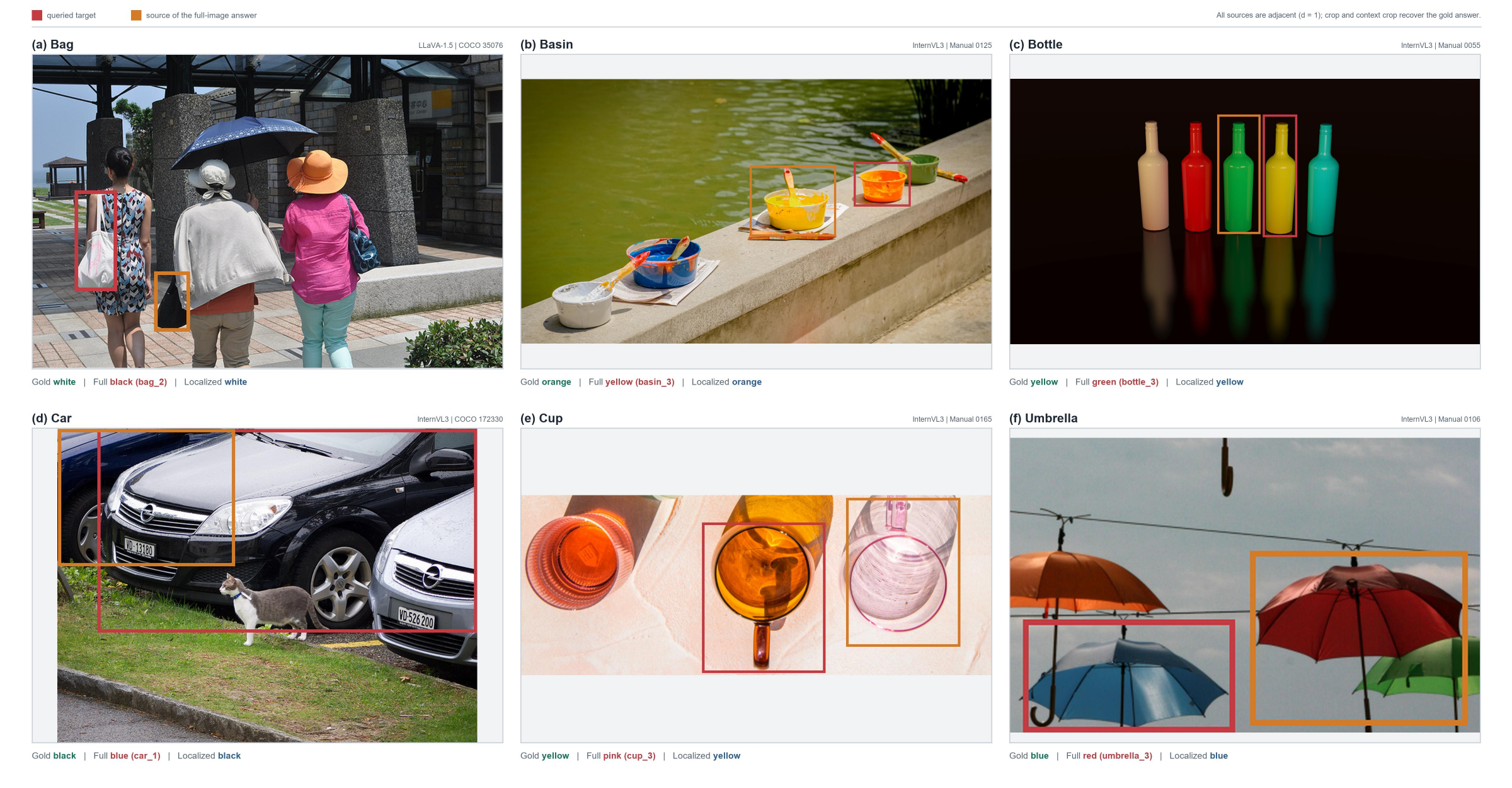}
\caption{Six manually selected, intervention-confirmed adjacent misbindings. The queried target is outlined in red and the source of the full-image answer in orange. In every panel, the full-image prediction is a visible attribute of the adjacent source instance, whereas both crop and context-crop evaluation recover the gold answer. The first case is from LLaVA-1.5 and the remaining cases are from InternVL3.}
\label{fig:qualitative_cases}
\end{figure*}

Figure~\ref{fig:qualitative_cases} shows six adjacent transfers across bags, basins, bottles, cars, cups, and umbrellas. Each full-image prediction is the attribute of the annotated distance-1 source, while crop and context-crop recover the target value. An identifiable adjacent source plus localized recovery is more consistent with same-class competition than arbitrary color guessing, although the joint image-and-query intervention does not establish an internal causal mechanism. Complete questions, ordered attribute sequences, and outputs are included in the release metadata.

Such transfers can return the wrong belonging in assistive search, select the wrong inventory or robot-picking item, or retrieve the wrong vehicle in traffic analysis. These are risk mappings rather than measured deployment outcomes, but they show why an in-image answer is not necessarily safe: object-presence checks cannot flag a wrong owner when every class and color is visible.

\section{Intervention Analysis}

\subsection{Protocol}
Interventions are applied to the 4261 L1/L3 questions with a single defined target and attribute. All intervention images preserve aspect ratio. A source image is downsampled only when its longer side exceeds 1800 pixels, with the target box scaled by the same factor. Crop oracle uses the annotated target box with horizontal and vertical padding equal to 15\% of target width and height, clipped to image boundaries. Context crop uses 50\% padding to retain more local context. Because an original expression such as ``third from the left'' is no longer meaningful after cropping, the direct crop prompt asks for the depicted instance's attribute and the context prompt asks for the attribute of the instance closest to the crop center.

``Oracle'' denotes access to the ground-truth target box, not perfect pixel isolation. A 15\% crop can retain fragments of an overlapping neighbor, while the 50\% context crop intentionally preserves nearby entities and scene cues. The comparison is therefore a joint visual-and-query localization test: it asks whether making the target easier to isolate reduces source-attributable misbinding.

\begin{table*}[t]
\centering
\caption{Intervention results on the L1/L3 subset. Crop and context crops are compared against full images.}
\label{tab:intervention}
\small
\begin{tabular}{lrrrrrr}
\toprule
Model & Full Acc. & Full MBR & Crop Acc. & Crop MBR & Ctx Acc. & Ctx MBR \\
\midrule
Qwen2.5-VL-7B & 69.56 & 14.57 & 69.44 & 12.72 & 73.36 & 11.24 \\
InternVL3-8B & 69.87 & 13.59 & 66.7 & 13.42 & 70.52 & 12.27 \\
LLaVA-1.5-7B & 47.92 & 34.33 & 68.06 & 15.21 & 66.86 & 18.21 \\
LLaVA-OneVision-7B & 63.98 & 20.56 & 69.56 & 15.61 & 69.54 & 15.84 \\
MiniCPM-V-2.6 & 62.22 & 21.76 & 71.44 & 11.19 & 72.8 & 11.92 \\
Gemini-3.5-Flash & 71.46 & 7.16 & 71.53 & 8.97 & 70.55 & 9.18 \\
Qwen3-VL-Plus & 74.23 & 8.89 & 71.84 & 10.02 & 72.64 & 11.24 \\
\bottomrule
\end{tabular}
\end{table*}


Table~\ref{tab:intervention} separates models whose errors are sensitive to visual competition from those that need broader context. LLaVA-1.5 gains 20.14 accuracy points and reduces MBR by 19.12 points under crop oracle; MiniCPM-V-2.6 gains 9.22 accuracy points and reduces MBR by 10.57 points. LLaVA-OneVision also reduces MBR by 4.95 points. These paired changes are stronger evidence of competition-sensitive binding than accuracy alone: the intervention removes competing instances, and the specific error component attributed to those instances falls.

The response is not universal. Gemini-3.5-Flash and Qwen3-VL-Plus begin with lower full-image MBR but increase under localized views. Proprietary preprocessing prevents separation of global-context removal from crop-induced scale or distribution shift. The negative gaps therefore do not imply that visual competition is beneficial; they show that target isolation is not a universally safe mitigation. This model-dependent pattern is another diagnostic advantage of \dataset: a proposed intervention can be evaluated on the failure type it is intended to repair, rather than judged only by aggregate accuracy.

\begin{table}[t]
\centering
\caption{Image-cluster bootstrap 95\% confidence intervals for MBR reduction on L1/L3. Positive values indicate that the intervention reduces MBR relative to full-image evaluation.}
\label{tab:intervention_ci}
\scriptsize
\renewcommand{\arraystretch}{1.12}
\begin{tabular}{lcc}
\toprule
Model & Crop & Context \\
\midrule
Qwen2.5-VL & +1.85 [-0.12, +3.69] & +3.33 [+1.38, +5.18] \\
InternVL3 & +0.16 [-1.68, +2.06] & +1.31 [-0.38, +2.93] \\
LLaVA-1.5 & +19.13 [+16.73, +21.41] & +16.12 [+13.62, +18.40] \\
LLaVA-OV & +4.95 [+2.89, +6.94] & +4.72 [+2.73, +6.64] \\
MiniCPM-V & +10.56 [+8.64, +12.45] & +9.83 [+7.87, +11.78] \\
Gemini & -1.81 [-3.18, -0.23] & -2.02 [-3.52, -0.54] \\
Qwen3-VL+ & -1.13 [-2.74, +0.35] & -2.35 [-3.87, -0.88] \\
\bottomrule
\end{tabular}
\end{table}


The image-cluster intervals in Table~\ref{tab:intervention_ci} reinforce this interpretation. MBR reduction is reliably positive for LLaVA-1.5, LLaVA-OneVision, and MiniCPM-V, while Qwen2.5-VL's context crop is positive but its crop interval crosses zero. InternVL3's intervals also cross zero, and the API-model intervals are zero-crossing or negative. Localization therefore identifies a substantial same-class competition component in selected open-source models, but not a single mechanism shared equally by every system.

\begin{figure*}[t]
    \centering
    \includegraphics[width=\textwidth]{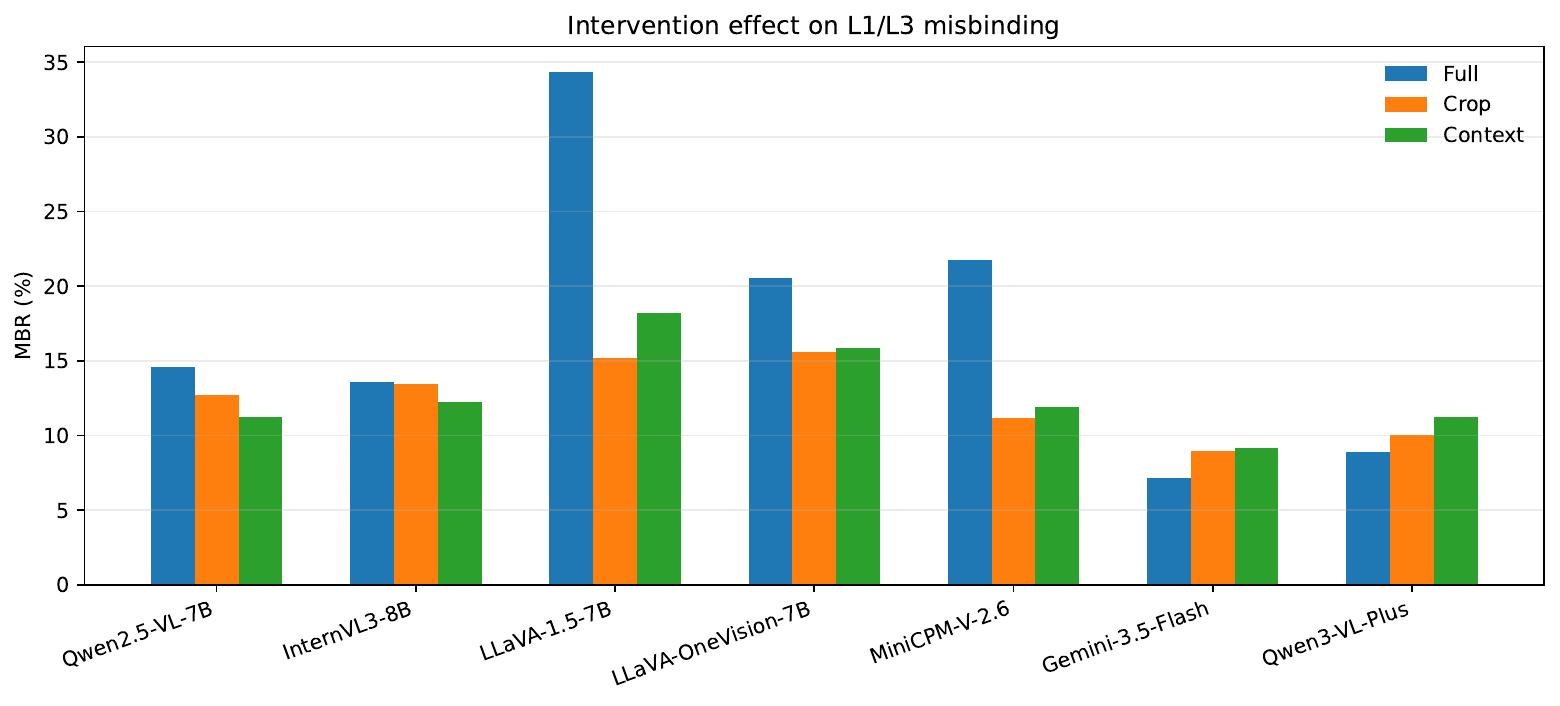}
    \caption{Intervention effect on L1/L3 MBR. Localization sharply reduces source-attributable misbinding for several open-source models but can remove useful context or shift the input distribution for stronger API systems.}
    \label{fig:intervention_mbr}
\end{figure*}

\subsection{Lightweight Inference-Time Mitigation}
\begin{table*}[t]
\centering
\caption{Instance-first prompting on open-source LVLMs.}
\label{tab:instance_first}
\small
\begin{tabular}{lrrrrrr}
\toprule
Model & Full Acc. & IF Acc. & Acc. Gap & Full MBR & IF MBR & MBR Gap \\
\midrule
Qwen2.5-VL-7B & 75.89 & 73.19 & -2.69 & 13.65 & 12.76 & -0.9 \\
InternVL3-8B & 76.69 & 78.24 & 1.54 & 14.01 & 9.65 & -4.36 \\
LLaVA-1.5-7B & 54.37 & 31.8 & -22.58 & 34.01 & 35.16 & 1.15 \\
LLaVA-OneVision-7B & 72.3 & 70.66 & -1.64 & 18.58 & 19.83 & 1.25 \\
MiniCPM-V-2.6 & 71.18 & 73.35 & 2.17 & 18.96 & 15.5 & -3.46 \\
\bottomrule
\end{tabular}
\end{table*}


Instance-first prompting explicitly externalizes an intermediate binding map: the model first enumerates same-class entities from left to right with their attributes and then answers the original question. No training or architectural access is required. InternVL3 improves accuracy by 1.54 points while reducing MBR by 4.36 points, and MiniCPM-V improves by 2.17 points while reducing MBR by 3.46 points. These paired gains indicate that explicit decomposition can repair a subset of binding errors.

Other models expose a different trade-off. Qwen2.5-VL reduces MBR slightly while losing accuracy, whereas LLaVA-1.5 loses 22.58 accuracy points and increases MBR under the longer prompt. A standard score would report only improvement or degradation; the binding-specific evaluation reveals whether the change actually repairs wrong-instance transfer. The mixed outcome positions \dataset as a testbed for future grounding-aware decoding, contrastive neighbor suppression, or targeted fine-tuning: mitigation should lower MBR without sacrificing general recognition and instruction following.

\section{Discussion}

\subsection{Why a Source-Aware Benchmark Is Necessary}
Standard accuracy answers whether a response is correct; object-hallucination measures answer whether predicted content is supported; grounding measures answer whether a referent is localized. None of these quantities alone answers whether a visible attribute was assigned to its owner. \dataset adds that missing variable by recording target and competing source instances. The resulting decomposition is operationally important: unsupported generation calls for hallucination suppression, recognition failure calls for stronger visual features, and adjacent transfer calls for better instance separation or binding. Collapsing them into one error rate obscures which component a new model or mitigation actually improves.

The seven-model results illustrate the consequence. API models lead aggregate accuracy, yet retain measurable MBR and high A-MBR. Six of seven models produce wrong-instance transfers more often than out-of-set hallucinations. A model can therefore appear strong under broad capability and object-presence tests while remaining unreliable for entity-specific questions. \dataset does not replace those benchmarks; it prevents their scores from being interpreted as evidence of instance-grounded reliability without a direct binding test.

\subsection{Adjacent Structure and Mechanistic Implications}
Across architectures, most identifiable transfers originate from ordinal distance 1. The bootstrap analysis shows that this neighbor concentration persists under image-level resampling even when total MBR differs substantially. This result supports local same-class competition rather than uniform answer noise and motivates mechanisms that explicitly contrast adjacent instances. The current ordinal analysis does not establish a causal effect of Euclidean separation: adjacent indices can have unequal pixel gaps, object scales, and overlaps. Normalized center distance, boundary distance, and overlap are therefore required to distinguish order-local competition from physical proximity.

\subsection{Implications for Multimedia Systems}
The cases in Fig.~\ref{fig:qualitative_cases} map this distinction to multimedia operations: attributes cue assistive identification, media and catalog search, robotic selection, and traffic retrieval. Real visual evidence can still identify the wrong entity, producing wrong-item retrieval or action rather than merely an awkward caption. Source-aware evaluation is therefore required before treating an LVLM as reliable for entity-specific decisions.

\subsection{Diagnostic Scope and Mitigation}
The controlled color domain enables deterministic source attribution and exposes a clean binding signal. It does not imply that actions, materials, states, or part attributes will exhibit identical rates. Those semantics introduce temporal and annotation ambiguity and require adapted question templates and judges. Similarly, crop and instance-first prompting are probes rather than a complete solution. Their model-dependent effects point toward two research directions: binding-aware training that contrasts neighboring same-class instances, and inference methods that preserve global context while explicitly grounding the target. The benchmark provides the boxes, neighbor structure, and error labels needed to evaluate whether such methods repair binding rather than merely shift aggregate accuracy.

\section{Data Availability and Ethical Considerations}

\dataset is intended to be released with annotations, generated questions, evaluation scripts, parser code, and model-output metadata needed to reproduce the reported tables and figures. For images originating from established datasets, release procedures will follow the original terms and provide source identifiers when direct redistribution is not appropriate. For manually collected web images, redistribution will be limited to files with recoverable license evidence permitting such use; otherwise, the public package will provide derived annotations and exclude the image file. Self-shot images can be released directly by the authors.

The benchmark contains images of people only for visual attribute binding evaluation. Identity, demographic, and sensitive-attribute questions are excluded; person questions concern only visible upper-clothing color. Manual review excludes potentially offensive, private, or ambiguous depictions. The dataset is designed for diagnostic evaluation of LVLM behavior and should not be used for surveillance, identity inference, or demographic profiling.

\section{Conclusion}

This study formalizes \problem as a source-identifiable failure in which an LVLM predicts an attribute present in the image but assigns it to the wrong same-class instance. \dataset addresses a gap left by aggregate VQA, object-hallucination, grounding, and compositionality evaluation: its controlled groups, ordered instance annotations, and four question levels reveal not only that an answer is wrong, but whether its evidence was copied from a specific visible neighbor.

Evaluation across seven LVLMs shows that instance-level reliability cannot be inferred from general accuracy alone. Open-source models average 19.84\% MBR and API models 7.55\%, while both groups retain approximately 81\% adjacent concentration among identifiable transfers. Image-cluster confidence intervals, a stratified parser audit, real-image cases, and controlled localization and prompting interventions support the same conclusion: DSCAM is a reproducible, spatially structured error category, and mitigation must reduce wrong-instance transfer without discarding useful context or general capability.

\subsection{Future Work}
Future work will turn \dataset into a controlled model-development loop. Leakage-safe train, validation, and held-out test splits will support comparisons of box-conditioned visual tokens, coordinate embeddings, neighbor-contrastive losses, and ordered instance--attribute supervision. Instance-first prompting and neighbor-contrastive decoding will remain inference-time baselines. All variants will use the same frozen test set and parser; success requires lower MBR and Err-MBR, stable or higher accuracy, and no increase in out-of-set hallucination or invalid answers. A claimed spatial repair must also reduce A-MBR or Distance-MBR, with image-cluster bootstrap intervals, paired image-level tests, and stratification by class, level, group size, position, and source distance.

Expansion will balance underrepresented classes and extend attributes to material, state, action, part, depth, and temporal identity using auditable evaluation. Ordinal distance will be complemented by normalized pixel distance, overlap, depth, and occlusion; parser audits will include API outputs and broader strata. Successful mitigations will finally be tested on newly collected and cross-dataset images to establish transfer beyond the controlled color slice.


\bibliographystyle{IEEEtran}
\bibliography{refs}

@inproceedings{rohrbach2018object,
  title={Object Hallucination in Image Captioning},
  author={Rohrbach, Anna and Hendricks, Lisa Anne and Burns, Kaylee and Darrell, Trevor and Saenko, Kate},
  booktitle={Proceedings of the 2018 Conference on Empirical Methods in Natural Language Processing},
  pages={4035--4045},
  year={2018},
  address={Brussels, Belgium},
  publisher={Association for Computational Linguistics},
  doi={10.18653/v1/D18-1437}
}

@inproceedings{li2023pope,
  title={Evaluating Object Hallucination in Large Vision-Language Models},
  author={Li, Yifan and Du, Yifan and Zhou, Kun and Wang, Jinpeng and Zhao, Xin and Wen, Ji-Rong},
  booktitle={Proceedings of the 2023 Conference on Empirical Methods in Natural Language Processing},
  pages={292--305},
  year={2023},
  address={Singapore},
  publisher={Association for Computational Linguistics},
  doi={10.18653/v1/2023.emnlp-main.20}
}

@article{wang2023amber,
  title={AMBER: An LLM-free Multi-dimensional Benchmark for MLLMs Hallucination Evaluation},
  author={Wang, Junyang and Wang, Yuhang and Xu, Guohai and Zhang, Jing and Gu, Yukai and Jia, Haitao and Wang, Jiaqi and Xu, Haiyang and Yan, Ming and Zhang, Ji and Sang, Jitao},
  journal={arXiv preprint arXiv:2311.07397},
  year={2023}
}

@inproceedings{guan2024hallusionbench,
  title={HallusionBench: An Advanced Diagnostic Suite for Entangled Language Hallucination and Visual Illusion in Large Vision-Language Models},
  author={Guan, Tianrui and Liu, Fuxiao and Wu, Xiyang and Xian, Ruiqi and Li, Zongxia and Liu, Xiaoyu and Wang, Xijun and Chen, Lichang and Huang, Furong and Yacoob, Yaser and Manocha, Dinesh and Zhou, Tianyi},
  booktitle={Proceedings of the IEEE/CVF Conference on Computer Vision and Pattern Recognition},
  year={2024}
}

@inproceedings{antol2015vqa,
  title={VQA: Visual Question Answering},
  author={Antol, Stanislaw and Agrawal, Aishwarya and Lu, Jiasen and Mitchell, Margaret and Batra, Dhruv and Zitnick, C. Lawrence and Parikh, Devi},
  booktitle={Proceedings of the IEEE International Conference on Computer Vision},
  year={2015}
}

@inproceedings{lin2014coco,
  title={Microsoft COCO: Common Objects in Context},
  author={Lin, Tsung-Yi and Maire, Michael and Belongie, Serge and Hays, James and Perona, Pietro and Ramanan, Deva and Doll{\'a}r, Piotr and Zitnick, C. Lawrence},
  booktitle={Proceedings of the European Conference on Computer Vision},
  pages={740--755},
  year={2014}
}

@inproceedings{hudson2019gqa,
  title={GQA: A New Dataset for Real-World Visual Reasoning and Compositional Question Answering},
  author={Hudson, Drew A. and Manning, Christopher D.},
  booktitle={Proceedings of the IEEE/CVF Conference on Computer Vision and Pattern Recognition},
  pages={6700--6709},
  year={2019}
}

@article{krishna2017visualgenome,
  title={Visual Genome: Connecting Language and Vision Using Crowdsourced Dense Image Annotations},
  author={Krishna, Ranjay and Zhu, Yuke and Groth, Oliver and Johnson, Justin and Hata, Kenji and Kravitz, Joshua and Chen, Stephanie and Kalantidis, Yannis and Li, Li-Jia and Shamma, David A. and Bernstein, Michael S. and Fei-Fei, Li},
  journal={International Journal of Computer Vision},
  volume={123},
  number={1},
  pages={32--73},
  year={2017},
  doi={10.1007/s11263-016-0981-7}
}

@inproceedings{pham2021vaw,
  title={Learning To Predict Visual Attributes in the Wild},
  author={Pham, Khoi and Kafle, Kushal and Lin, Zhe and Ding, Zhihong and Cohen, Scott and Tran, Quan and Shrivastava, Abhinav},
  booktitle={Proceedings of the IEEE/CVF Conference on Computer Vision and Pattern Recognition},
  pages={13018--13028},
  year={2021}
}

@inproceedings{johnson2017clevr,
  title={CLEVR: A Diagnostic Dataset for Compositional Language and Elementary Visual Reasoning},
  author={Johnson, Justin and Hariharan, Bharath and van der Maaten, Laurens and Fei-Fei, Li and Zitnick, C. Lawrence and Girshick, Ross},
  booktitle={Proceedings of the IEEE Conference on Computer Vision and Pattern Recognition},
  year={2017}
}

@inproceedings{thrush2022winoground,
  title={Winoground: Probing Vision and Language Models for Visio-Linguistic Compositionality},
  author={Thrush, Tristan and Jiang, Ryan and Bartolo, Max and Singh, Amanpreet and Williams, Adina and Kiela, Douwe and Ross, Candace},
  booktitle={Proceedings of the IEEE/CVF Conference on Computer Vision and Pattern Recognition},
  pages={5238--5248},
  year={2022}
}

@inproceedings{yuksekgonul2023bags,
  title={When and Why Vision-Language Models Behave like Bags-Of-Words, and What to Do About It?},
  author={Yuksekgonul, Mert and Bianchi, Federico and Kalluri, Pratyusha and Jurafsky, Dan and Zou, James},
  booktitle={International Conference on Learning Representations},
  year={2023}
}

@inproceedings{liu2023llava,
  title={Visual Instruction Tuning},
  author={Liu, Haotian and Li, Chunyuan and Wu, Qingyang and Lee, Yong Jae},
  booktitle={Advances in Neural Information Processing Systems},
  volume={36},
  pages={34892--34916},
  year={2023}
}

@article{liu2023improvedllava,
  title={Improved Baselines with Visual Instruction Tuning},
  author={Liu, Haotian and Li, Chunyuan and Li, Yuheng and Lee, Yong Jae},
  journal={arXiv preprint arXiv:2310.03744},
  year={2023}
}

@article{li2024llavaonevision,
  title={LLaVA-OneVision: Easy Visual Task Transfer},
  author={Li, Bo and Zhang, Yuanhan and Guo, Dong and Zhang, Renrui and Li, Feng and Zhang, Hao and Zhang, Kaichen and Zhang, Peiyuan and Li, Yanwei and Liu, Ziwei and Li, Chunyuan},
  journal={arXiv preprint arXiv:2408.03326},
  year={2024}
}

@article{bai2025qwen25vl,
  title={Qwen2.5-VL Technical Report},
  author={Bai, Shuai and Chen, Keqin and Liu, Xuejing and Wang, Jialin and Ge, Wenbin and Song, Sibo and Dang, Kai and Wang, Peng and Wang, Shijie and Tang, Jun and Zhong, Humen and Zhu, Yuanzhi and Yang, Ming-Hsuan and Li, Zhaohai and Wan, Jianqiang and Wang, Pengfei and Ding, Wei and Fu, Zheren and Xu, Yiheng and Ye, Jiabo and Zhang, Xi and Xie, Tianbao and Cheng, Zesen and Zhang, Hang and Yang, Zhibo and Xu, Haiyang and Lin, Junyang},
  journal={arXiv preprint arXiv:2502.13923},
  year={2025}
}

@article{zhu2025internvl3,
  title={InternVL3: Exploring Advanced Training and Test-Time Recipes for Open-Source Multimodal Models},
  author={Zhu, Jinguo and Wang, Weiyun and Chen, Zhe and Liu, Zhaoyang and Ye, Shenglong and Gu, Lixin and Tian, Hao and Duan, Yuchen and Su, Weijie and Shao, Jie and Gao, Zhangwei and Cui, Erfei and Wang, Xuehui and Cao, Yue and Liu, Yangzhou and Wei, Xingguang and Zhang, Hongjie and Wang, Haomin and Xu, Weiye and Li, Hao and Wang, Jiahao and Deng, Nianchen and Li, Songze and He, Yinan and Jiang, Tan and Luo, Jiapeng and Wang, Yi and He, Conghui and Shi, Botian and Zhang, Xingcheng and Shao, Wenqi and He, Junjun and Xiong, Yingtong and Qu, Wenwen and Sun, Peng and Jiao, Penglong and Lv, Han and Wu, Lijun and Zhang, Kaipeng and Deng, Huipeng and Ge, Jiaye and Chen, Kai and Wang, Limin and Dou, Min and Lu, Lewei and Zhu, Xizhou and Lu, Tong and Lin, Dahua and Qiao, Yu and Dai, Jifeng and Wang, Wenhai},
  journal={arXiv preprint arXiv:2504.10479},
  year={2025}
}

@article{yao2024minicpmv,
  title={MiniCPM-V: A GPT-4V Level MLLM on Your Phone},
  author={Yao, Yuan and Yu, Tianyu and Zhang, Ao and Wang, Chongyi and Cui, Junbo and Zhu, Hongji and Cai, Tianchi and Li, Haoyu and Zhao, Weilin and He, Zhihui and others},
  journal={arXiv preprint arXiv:2408.01800},
  year={2024}
}

@misc{google2026gemini35flash,
  title={Gemini 3.5 Flash Model Card},
  author={{Google DeepMind}},
  year={2026},
  howpublished={\url{https://deepmind.google/models/model-cards/gemini-3-5-flash/}},
  note={Accessed: 2026-07-17}
}

@misc{qwencloud2026qwen3vlplus,
  title={Qwen3-VL-Plus},
  author={{Qwen Cloud}},
  year={2026},
  howpublished={\url{https://www.qwencloud.com/models/qwen3-vl-plus}},
  note={Accessed: 2026-07-17}
}

@inproceedings{goyal2017vqav2,
  title={Making the V in VQA Matter: Elevating the Role of Image Understanding in Visual Question Answering},
  author={Goyal, Yash and Khot, Tejas and Summers-Stay, Douglas and Batra, Dhruv and Parikh, Devi},
  booktitle={Proceedings of the IEEE Conference on Computer Vision and Pattern Recognition},
  pages={6904--6913},
  year={2017}
}

@inproceedings{zhu2016visual7w,
  title={Visual7W: Grounded Question Answering in Images},
  author={Zhu, Yuke and Groth, Oliver and Bernstein, Michael and Fei-Fei, Li},
  booktitle={Proceedings of the IEEE Conference on Computer Vision and Pattern Recognition},
  pages={4995--5004},
  year={2016}
}

@article{plummer2017flickr30k,
  title={Flickr30K Entities: Collecting Region-to-Phrase Correspondences for Richer Image-to-Sentence Models},
  author={Plummer, Bryan A. and Wang, Liwei and Cervantes, Christopher M. and Caicedo, Juan C. and Hockenmaier, Julia and Lazebnik, Svetlana},
  journal={International Journal of Computer Vision},
  volume={123},
  number={1},
  pages={74--93},
  year={2017}
}

@inproceedings{kazemzadeh2014referitgame,
  title={ReferItGame: Referring to Objects in Photographs of Natural Scenes},
  author={Kazemzadeh, Sahar and Ordonez, Vicente and Matten, Mark and Berg, Tamara},
  booktitle={Proceedings of the 2014 Conference on Empirical Methods in Natural Language Processing},
  pages={787--798},
  year={2014},
  address={Doha, Qatar},
  publisher={Association for Computational Linguistics},
  doi={10.3115/v1/D14-1086}
}

@inproceedings{yu2016referringcontext,
  title={Modeling Context in Referring Expressions},
  author={Yu, Licheng and Poirson, Patrick and Yang, Shan and Berg, Alexander C. and Berg, Tamara L.},
  booktitle={Proceedings of the European Conference on Computer Vision},
  pages={69--85},
  year={2016}
}

@inproceedings{yu2018mattnet,
  title={MAttNet: Modular Attention Network for Referring Expression Comprehension},
  author={Yu, Licheng and Lin, Zhe and Shen, Xiaohui and Yang, Jimei and Lu, Xin and Bansal, Mohit and Berg, Tamara L.},
  booktitle={Proceedings of the IEEE Conference on Computer Vision and Pattern Recognition},
  pages={1307--1315},
  year={2018}
}

@inproceedings{zhao2022vlchecklist,
  title={An Explainable Toolbox for Evaluating Pre-trained Vision-Language Models},
  author={Zhao, Tiancheng and Zhang, Tianqi and Zhu, Mingwei and Shen, Haozhan and Lee, Kyusong and Lu, Xiaopeng and Yin, Jianwei},
  booktitle={Proceedings of the 2022 Conference on Empirical Methods in Natural Language Processing: System Demonstrations},
  pages={30--37},
  year={2022},
  address={Abu Dhabi, UAE},
  publisher={Association for Computational Linguistics},
  doi={10.18653/v1/2022.emnlp-demos.4}
}

@inproceedings{ma2023crepe,
  title={CREPE: Can Vision-Language Foundation Models Reason Compositionally?},
  author={Ma, Zixian and Hong, Jerry and Gul, Mustafa Omer and Gandhi, Mona and Gao, Irena and Krishna, Ranjay},
  booktitle={Proceedings of the IEEE/CVF Conference on Computer Vision and Pattern Recognition},
  pages={10910--10921},
  year={2023}
}

@inproceedings{hsieh2023sugarcrepe,
  title={SugarCrepe: Fixing Hackable Benchmarks for Vision-Language Compositionality},
  author={Hsieh, Cheng-Yu and Zhang, Jieyu and Ma, Zixian and Kembhavi, Aniruddha and Krishna, Ranjay},
  booktitle={Advances in Neural Information Processing Systems},
  volume={36},
  year={2023}
}

@inproceedings{radford2021clip,
  title={Learning Transferable Visual Models From Natural Language Supervision},
  author={Radford, Alec and Kim, Jong Wook and Hallacy, Chris and Ramesh, Aditya and Goh, Gabriel and Agarwal, Sandhini and Sastry, Girish and Askell, Amanda and Mishkin, Pamela and Clark, Jack and Krueger, Gretchen and Sutskever, Ilya},
  booktitle={International Conference on Machine Learning},
  pages={8748--8763},
  year={2021}
}

@inproceedings{li2022blip,
  title={BLIP: Bootstrapping Language-Image Pre-training for Unified Vision-Language Understanding and Generation},
  author={Li, Junnan and Li, Dongxu and Xiong, Caiming and Hoi, Steven},
  booktitle={International Conference on Machine Learning},
  pages={12888--12900},
  year={2022}
}

@inproceedings{alayrac2022flamingo,
  title={Flamingo: A Visual Language Model for Few-Shot Learning},
  author={Alayrac, Jean-Baptiste and Donahue, Jeff and Luc, Pauline and Miech, Antoine and Barr, Iain and Hasson, Yana and Lenc, Karel and Mensch, Arthur and Millican, Katherine and Reynolds, Malcolm and others},
  booktitle={Advances in Neural Information Processing Systems},
  volume={35},
  pages={23716--23736},
  year={2022}
}

@inproceedings{li2023blip2,
  title={BLIP-2: Bootstrapping Language-Image Pre-training with Frozen Image Encoders and Large Language Models},
  author={Li, Junnan and Li, Dongxu and Savarese, Silvio and Hoi, Steven},
  booktitle={International Conference on Machine Learning},
  year={2023}
}

@inproceedings{dai2023instructblip,
  title={InstructBLIP: Towards General-purpose Vision-Language Models with Instruction Tuning},
  author={Dai, Wenliang and Li, Junnan and Li, Dongxu and Tiong, Anthony Meng Huat and Zhao, Junqi and Wang, Weisheng and Li, Boyang and Fung, Pascale and Hoi, Steven C. H.},
  booktitle={Advances in Neural Information Processing Systems},
  volume={36},
  year={2023}
}

@article{fu2023mme,
  title={MME: A Comprehensive Evaluation Benchmark for Multimodal Large Language Models},
  author={Fu, Chaoyou and Chen, Peixian and Shen, Yunhang and Qin, Yulei and Zhang, Mengdan and Lin, Xu and Qiu, Zhenyu and Lin, Wei and Yang, Jinrui and Zheng, Xiawu and Li, Ke and Sun, Xing and Ji, Rongrong},
  journal={arXiv preprint arXiv:2306.13394},
  year={2023}
}

@article{li2023seedbench,
  title={SEED-Bench: Benchmarking Multimodal LLMs with Generative Comprehension},
  author={Li, Bohao and Wang, Rui and Wang, Guangzhi and Ge, Yuying and Ge, Yixiao and Shan, Ying},
  journal={arXiv preprint arXiv:2307.16125},
  year={2023}
}

@inproceedings{yu2024mmvet,
  title={MM-Vet: Evaluating Large Multimodal Models for Integrated Capabilities},
  author={Yu, Weihao and Yang, Zhengyuan and Li, Linjie and Wang, Jianfeng and Lin, Kevin and Liu, Zicheng and Wang, Xinchao and Wang, Lijuan},
  booktitle={International Conference on Machine Learning},
  year={2024}
}

@inproceedings{liu2024mmbench,
  title={MMBench: Is Your Multi-modal Model an All-Around Player?},
  author={Liu, Yuan and Duan, Haodong and Zhang, Yuanhan and Li, Bo and Zhang, Songyang and Zhao, Wangbo and Yuan, Yike and Wang, Jiaqi and He, Conghui and Liu, Ziwei and Chen, Kai and Lin, Dahua},
  booktitle={Proceedings of the European Conference on Computer Vision},
  pages={216--233},
  year={2024}
}

@inproceedings{gunjal2024mhaldetect,
  title={Detecting and Preventing Hallucinations in Large Vision Language Models},
  author={Gunjal, Anisha and Yin, Jihan and Bas, Erhan},
  booktitle={Proceedings of the AAAI Conference on Artificial Intelligence},
  volume={38},
  number={16},
  pages={18135--18143},
  year={2024},
  doi={10.1609/aaai.v38i16.29771}
}

@inproceedings{huang2024opera,
  title={OPERA: Alleviating Hallucination in Multi-Modal Large Language Models via Over-Trust Penalty and Retrospection-Allocation},
  author={Huang, Qidong and Dong, Xiaoyi and Zhang, Pan and Wang, Bin and He, Conghui and Wang, Jiaqi and Lin, Dahua and Zhang, Weiming and Yu, Nenghai},
  booktitle={Proceedings of the IEEE/CVF Conference on Computer Vision and Pattern Recognition},
  pages={13418--13427},
  year={2024}
}

@inproceedings{saravanan2025binding,
  title={Investigating Mechanisms for In-Context Vision Language Binding},
  author={Saravanan, Darshana and Tapaswi, Makarand and Gandhi, Vineet},
  booktitle={2025 IEEE/CVF Conference on Computer Vision and Pattern Recognition Workshops (CVPRW)},
  pages={4852--4856},
  year={2025},
  doi={10.1109/CVPRW67362.2025.00476}
}

@article{cui2026dualbinding,
  title={The Dual Mechanisms of Spatial Variable Binding in Vision-Language Models},
  author={Cui, Kelly and Prakash, Nikhil and Messica, Shoval and Raina, Ayush and Bau, David and Torralba, Antonio and Shaham, Tamar Rott},
  journal={arXiv preprint arXiv:2603.22278},
  year={2026}
}

@article{tian2026multibind,
  title={MultiBind: A Benchmark for Attribute Misbinding in Multi-Subject Generation},
  author={Tian, Wenqing and Mao, Hanyi and Liu, Zhaocheng and Zhang, Lihua and Liu, Qiang and Wu, Jian and Wang, Liang},
  journal={arXiv preprint arXiv:2603.21937},
  year={2026}
}

@misc{guardian2018busadvert,
  author={{The Guardian}},
  title={Chinese Facial Recognition System Mistakes Bus Advert for Jaywalker},
  year={2018},
  howpublished={The Guardian, November 22, 2018},
  url={https://www.theguardian.com/world/2018/nov/22/chinese-facial-recognition-system-mistakes-bus-advert-for-jaywalker},
  note={Accessed: August 6, 2026}
}

@techreport{ntsb2019tempe,
  author={{National Transportation Safety Board}},
  title={Collision Between Vehicle Controlled by Developmental Automated Driving System and Pedestrian, Tempe, Arizona, March 18, 2018},
  institution={National Transportation Safety Board},
  number={HAR-19/03},
  year={2019},
  url={https://www.ntsb.gov/investigations/AccidentReports/Reports/HAR1903.pdf},
  note={Accessed: August 6, 2026}
}

\end{document}